\documentclass[11pt,a4paper]{article}

\usepackage{times}
\usepackage{latexsym}
\usepackage{amsmath}
\usepackage{graphicx}
\usepackage{authblk}
\usepackage[hyperref]{acl2017}
\usepackage{url}
\aclfinalcopy 
\def\aclpaperid{195} 

\newcommand{\Ni}{({\em i})~}
\newcommand{\Nii}{({\em ii})~}

\usepackage{color}

\title{A Context-Aware Approach\\
for Detecting Check-Worthy Claims in Political Debates}
\author[1]{Pepa Gencheva}
\author[1]{Ivan Koychev}
\author[2]{Llu\'{i}s M\`{a}rquez}
\author[2]{Alberto Barr\'on-Cede\~no}
\author[2]{Preslav Nakov}

\affil[1]{Sofia University ``St. Kliment Ohridski'', Bulgaria}
\affil[2]{Qatar Computing Research Institute, HBKU, Qatar}
\affil[ ]{\textit{pepa.k.gencheva@gmail.com}, 
\textit{koychev@fmi.uni-sofia.bg}}
\affil[ ]{\textit{\{pnakov, lmarquez, albarron\}@hbku.edu.qa}}

\date{September 2017}
\begin{document}
\maketitle
\begin{abstract}

In the context of investigative journalism, we address the problem of automatically identifying which claims in a given document are most worthy and should be prioritized for fact-checking. Despite its importance, this is a relatively understudied problem. Thus, we create a new dataset of political debates, containing statements that have been fact-checked by nine reputable sources, and we train machine learning models to predict which claims should be prioritized for fact-checking, i.e., we model the problem as a ranking task. 
Unlike previous work, which has looked primarily at sentences in isolation, in this paper we focus on a rich input representation modeling the context: relationship between the target statement and the larger context of the debate, interaction between the opponents, and reaction by the moderator and by the public. Our experiments show state-of-the-art results, outperforming a strong rivaling system by a margin, while also confirming the importance of the contextual information.

\end{abstract}

\section{Introduction}

The current coverage of the political landscape in the press and in social media has led to an unprecedented situation. Like never before, a statement in an interview, a press release, a blog note, or a tweet can spread almost instantaneously
and reach the public in no time. 
This proliferation speed has left little time for double-checking claims against the facts, which has proven critical in politics, e.g.,~during the 2016 presidential campaign in the USA, which was arguably impacted by fake news in social media and by false claims.

\begin{figure}
\footnotesize
\centering
\includegraphics[width=230px]{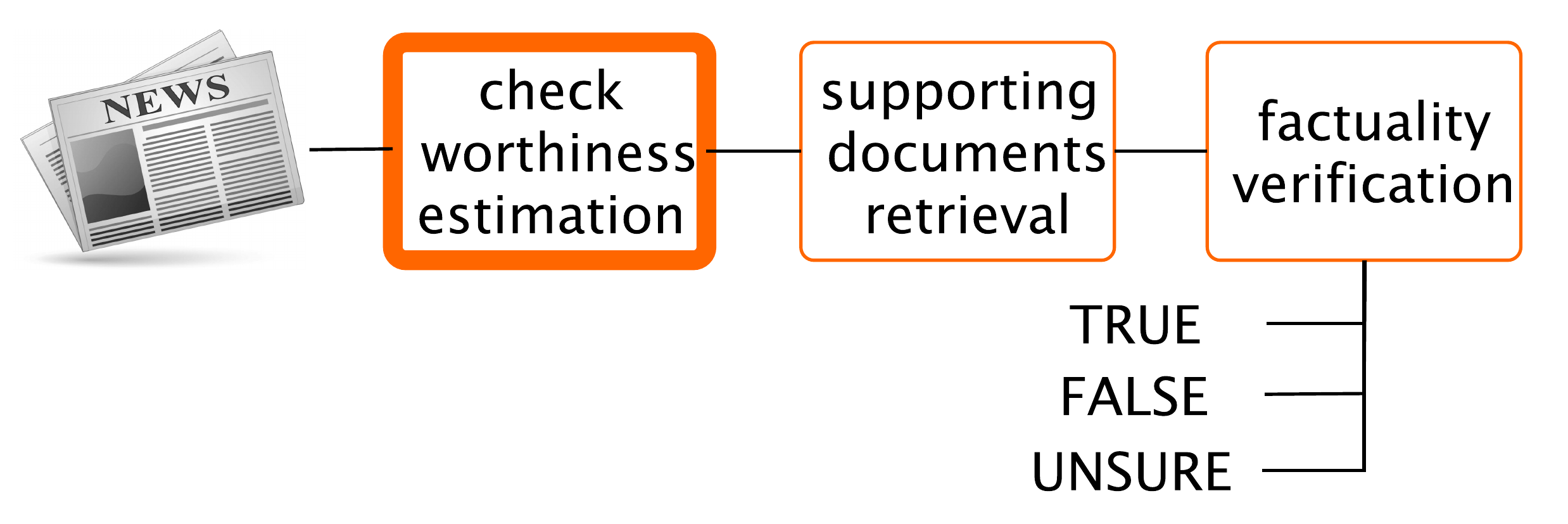}
\caption{Information verification pipeline.}
\label{fig:pipeline}
\end{figure}

\noindent Investigative journalists and volunteers have been working hard trying to get to the root of a claim and to present solid evidence in favor or against it. Manual fact-checking has proven very time-consuming, and thus automatic methods have been proposed as a way to speed-up the process. 
For instance, there has been  work on checking the factuality/credibility of a claim, of a news article, or of an information source~\cite{castillo2011information,ba2016vera, zubiaga2016analysing, ma2016detecting, Hardalov2016,RANLP2017:clickbait,RANLP2017:factchecking:external,RANLP2017:credibility:trolls}. However, less attention has been paid to other steps of the fact-checking pipeline, which is shown in Figure~\ref{fig:pipeline}. 

The process starts when a document is made public. First, an intrinsic analysis is carried out in which check-worthy text fragments are identified.
Then, other documents that might support or rebut a claim in the document are retrieved from various sources. Finally, by comparing a claim against the retrieved evidence, a system can determine whether the claim is likely true or likely false. For instance, \citet{10.1371journal.pone.0128193} do this on the basis of a knowledge graph derived from Wikipedia. The outcome could then be presented to a human expert for final judgment.\footnote{As of present, fully automatic methods for fact checking still lag behind in terms of quality, and thus also of credibility in the eyes of the users, compared to what high-quality manual checking by reputable sources can achieve, which means that a final double-checking by a human expert is needed.}

\noindent In this paper, we focus on the first step: predicting check-worthiness of claims. Our contributions can be summarized as follows:
\begin{enumerate}
\item \emph{New dataset:} We build a new dataset of manually-annotated claims, extracted from the 2016 US presidential and vice-presidential debates, which we gathered from nine reputable sources such as CNN, NPR, and PolitiFact, and which we release to the research community.
\item \emph{Modeling the context:}  We develop a novel approach for automatically predicting which claims should be prioritized for fact-checking, based on a rich input representation. In particular, we model not only the textual content, but also the context: how the target claim relates to the current segment, to neighboring segments and sentences, and to the debate as a whole, and also how the opponents and the public react to it. 
\item {State-of-the-art results:} We achieve state-of-the-art results, outperforming a strong rivaling system by a margin, while also demonstrating that this improvement is due primarily to our modeling of the context.
\end{enumerate}
We model the problem as a  ranking task, and we train both Support Vector Machines (SVM) and Feed-forward Neural Networks (FNN) obtaining state-of-the-art results. We also analyze the relevance of the specific feature groups and we show that modeling the context yields a significant boost in performance. Finally, we also analyze 
whether we can learn to predict which facts are check-worthy with respect to each of the individual media sources, thus capturing their biases.
It is worth noting that while trained on political debates, many features of our model can be potentially applied to other kinds of information sources, e.g., interviews and news.

The rest of the paper is organized as follows: Section~\ref{sec:related} discusses related work. Section~\ref{sec:data} describes the process of gathering and annotating our political debates dataset. Section~\ref{sec:model} presents our supervised approach to predicting fact-checking worthiness, including the explanation of the model and the information sources we use. Section~\ref{sec:evaluation} presents the evaluation setup and discusses the results.
Section~\ref{sec:discuss} provides further analysis.
Finally, Section~\ref{sec:conclusion} presents the conclusions and outlines some possible directions for future research.

\section{Related Work}
\label{sec:related}

The previous work that is most relevant to our work here is that of \cite{Hassan:15}, who developed the \emph{ClaimBuster} system, which assigns each sentence in a document a score, i.e., a number between 0 and 1 showing how worthy it is of fact-checking. The system is trained on their own dataset of about eight thousand debate sentences (1,673 of them check-worthy), annotated by students, university professors, and journalists. Unfortunately, this dataset is not publicly available and it contains sentences without context as about 60\% of the original sentences had to be thrown away due to lack of agreement. 

In contrast, we develop a new publicly-available dataset,\footnote{The dataset and the source code are available in GitHub: \url{https://github.com/pgencheva/claim-rank}}  based on manual annotations of political debates by nine highly-reputed fact-checking sources, where sentences are annotated in the context of the entire debate. This allows us to explore a novel approach, which focuses on the context.

Note also that the \emph{ClaimBuster} dataset is annotated following guidelines from \cite{Hassan:15} rather than a real fact-checking website; yet, it was evaluated against CNN and PolitiFact \cite{Hassan2016ComparingAF}. In contrast, we train and evaluate directly on annotations from fact-checking websites, and thus we learn to fit them better.

Beyond the document context, it has been proposed to mine check-worthy claims on the Web. For example, \citet{ennals2010disputed} searched for linguistic cues of disagreement between the author of a statement and what is believed, e.g., ``\emph{falsely claimed that X}''. 
The claims matching the patterns go through a statistical classifier, which marks the text of the claim. This procedure can be used to acquire a dataset of disputed claims from the Web.

Given a set of disputed claims, \cite{ennals2010highlighting} approached the task as locating new claims on the Web that entail the ones that have already been collected. Thus, the task can be conformed as recognizing textual entailment, which is analyzed in detail in \cite{dagan2009recognizing}.

Finally, \citet{le2016towards} argued that the top terms in claim vs.\ non-claim sentences are highly overlapping, which is a problem for bag-of-words approaches. Thus, they used a Convolutional Neural Network, where each word is represented by its embedding and each named entity is replaced by its tag, e.g., \emph{person}, \emph{organization}, \emph{location}. 

\begin{table}[tbh]
\small
\centering
\begin{tabular}{lrrrrr}
  \bf Medium & \bf 1st & \bf 2nd & \bf VP & \bf 3rd & \bf Total\\ 
 \hline
  ABC News& 35 & 50 & 29 & 28 & 142 \\
  Chicago Tribune & 30 & 29 & 31 & 38 & 128\\
  CNN & 46 & 30 & 37 & 60 & 173 \\
  FactCheck.org & 15 & 45 & 47 & 60 & 167 \\
  NPR & 99 & 92 & 91 & 89 & 371 \\
  PolitiFact & 74 & 62 & 60 & 57 & 253 \\
  The Guardian & 27 & 39 & 54 & 72 & 192 \\
  The New York Times & 26 & 25 & 46 & 52 & 149 \\
  The Washington Post & 26 & 19 & 33 & 17	 & 95 \\
 \hline
  \bf Total annotations & 378 & 391 & 428 & 473 &  1,670 \\
  \bf 
  Annotated sentences & 218 & 235 & 183 & 244 & 880 \\
\end{tabular}
\caption{Number of annotations in each medium for the 1st, 2nd and 3rd presidential and the vice-presidential debates.}
\label{table:per:medium}
\end{table}

\section{The CW-USPD-2016 dataset on US Presidential Debates}
\label{sec:data}

We created a new dataset called CW-USPD-2016 (check-worthiness in the US presidential debates 2016) for finding check-worthy claims in context.
In particular, we used four transcripts of the 2016 US election: one vice-presidential and three presidential debates. For each debate, we used the publicly-available manual analysis about it from nine reputable fact-checking sources, as shown in Table~\ref{table:per:medium}. This could include not just a statement about factuality, but any free text that journalists decided to add, e.g., links to biographies or behavioral analysis of the opponents and moderators. We converted this to binary annotation about whether a particular sentence was annotated for factuality by a given source.
Whenever one or more annotations were about part of a sentence, we selected the entire sentence, and when an annotation spanned over multiple sentences, we selected each of them.

Ultimately, we ended up with a dataset of four debates, with a total of 5,415 sentences.
The agreement between the sources was low as Table~\ref{table:agreement} shows: only one sentence was selected by all nine sources, 57 sentences by at least five, 197 by at least three, 388 by at least two, and 880 by at least one. The reason for this is that the different media aimed at annotating sentences according to their own editorial line, rather than trying to be exhaustive in any way. This suggests that the task of predicting which sentence would contain check-worthy claims will be challenging. Thus, below we focus on a ranking task rather than on absolute predictions. Moreover, we predict which sentence would be selected \Ni by at least one of the media, or \Nii by a specific medium.

\begin{table}[tbh]
\small
\centering
  \begin{tabular}{crr}
    \bf Agreement & \bf Number of & \bf Cumulative \\
    \bf Level & \bf Sentences & \bf Sum \\
    \hline
    9 & 1 & 1 \\
    8 & 6 & 7 \\
    7 & 5 & 12 \\
    6 & 19 & 31 \\
    5 & 26 & 57 \\
    4 & 40 & 97 \\
    3 & 100 & 197 \\
    2 & 191 & 388 \\
	\bf 1 & \bf 492 & \bf 880 \\
    \hline
    \multicolumn{3}{l}{\bf Total number of sentences: 5,415}\\
  \end{tabular}
\caption{Agreement between the media represented as the number of sentences that $n$ out of nine providers identified as check-worthy.}
\label{table:agreement}
\end{table}

\noindent Note that the investigative journalists did not select the check-worthy claims in isolation. Our analysis shows that these include claims that were highly disputed during the debate, that were relevant to the topic introduced by the moderator, etc. We will make use of these contextual dependencies below, which is something that was not previously tried in related work.

\section{Modeling Check-Worthiness}
\label{sec:model}

We developed a rich input representation in order to model and to learn the \emph{check-worthiness} concept. The feature types we implemented operate at the sentence- (S) and at the context-level (C), in either case targeting \emph{segments} by the same speaker.\footnote{We define a \emph{segment} as a maximal set of consecutive sentences by the same speaker without intervention by another speaker or by the moderator.}
The context features are novel and a contribution of this study. We also implemented a set of core features to compare to the state of the art. All of them are described below.

\subsection{Sentence-Level Features}

\textbf{ClaimBuster-based} (\emph{1,045 S features}; core): 
First, in order to be able to compare our model and features directly to the previous state of the art, we re-implemented, to the best of our ability, the sentence-level features of \emph{ClaimBuster} as described in ~\cite{Hassan:15}, namely TF.IDF-weighted bag of words (998 features), part-of-speech tags (25 features), named entities as recognized by \emph{Alchemy API}\footnote{\url{http://www.ibm.com/watson/alchemy-api.html}}
(20 features), sentiment score from Alchemy API (1~feature), and number of tokens in the target sentence (1 feature). 

\noindent Apart from providing means of comparison to the state of the art, these features also make a solid contribution to the final system we build for check-worthiness estimation. 
However, note that we did not have access to the training data of ClaimBuster, which is not publicly available, and we thus train on our dataset (described above). 

\textbf{Sentiment} (\emph{2 S features}): Some sentences are highly negative, which can signal the presence of an interesting claim to check, as the two example sentences below show (from the 1st and the 2nd presidential debates):

\vspace{3pt}
\begin{footnotesize}
\begin{tabular}{rp{55mm}}
Trump:	& Murders are up.	\\
Clinton:	&  Bullying is up.	\\
\\
\end{tabular}
\end{footnotesize}
We used the NRC sentiment lexicon~\cite{Mohammad13} as a source of words and $n$-grams with positive/negative sentiment, and we counted the number of positive and of negative words in the target sentence. These features are different from those in the \emph{CB features} above, where these lexicons were not used.

\textbf{Named entities (NE)} (\emph{1 S feature}): Sentences that contain named entity mentions are more likely to contain a claim that is worth fact-checking as they discuss particular people, organizations, and locations.
Thus, we have a feature that counts the number of named entity mentions in the target sentence; we use the \emph{NLTK toolkit} for named entity recognition~\cite{Loper02nltk:the}.
Unlike the \emph{CB features} above, here we only have one feature; we also use a different toolkit for named entity recognition.

\textbf{Linguistic features} (\emph{9 S features}): We count the number of words in each sentence that belong to each of the following lexicons: Language Bias lexicon \cite{Recasens+al:13a}, Opinion Negative and Positive Words \cite{Liu:2005:OOA:1060745.1060797}, Factives and Assertive Predicates \cite{hooper1974assertive}, Hedges \cite{hyland1998hedging}, Implicatives \cite{karttunen1971implicative}, and Strong and Weak subjective words. Some examples are shown in Table~\ref{table:linguistic}.

\begin{table}[htb]
\footnotesize
\centering
\begin{tabular}{ll}
\textbf{Feature Name} & \multicolumn{1}{c}{\textbf{Examples}} \\
\hline
Bias & capture, create, demand, follow\\ 
Negatives & abnormal, bankrupt, cheat, conflicts\\
Positives & accurate, achievements, affirm\\
Factives & realize, know, discover, learn\\
Assertives & think, believe, imagine, guarantee\\
Hedges & approximately, estimate, essentially\\
Implicatives & cause, manage, hesitate, neglect \\
Strong-subj & admire, afraid, agreeably, apologist\\
Weak-subj & abandon, adaptive, champ, consume\\
\hline
\end{tabular}
\caption{Linguistic features and examples.}
\label{table:linguistic}
\end{table}

\textbf{Tense} (\emph{1 S feature}): Most of the check-worthy claims mention past events. In order to detect when the speaker is making a reference to the past or s/he is talking about his/her future vision and plans, we include a feature with three values---indicating whether the text is in past, present, or future tense. The feature is extracted from the verbal expressions, using POS  tags and a list of auxiliary verbs and phrases such as \emph{will}, \emph{have to}, etc.

\textbf{Length} (\emph{1 S feature}): Shorter sentences are generally less likely to contain a check-worthy claim.\footnote{One notable exception are short sentences with negations, e.g.,~\emph{Wrong.}, \emph{Nonsense.}, etc.} 
Thus, we have a feature for the length of the sentence in terms of characters.
Note that this feature was not part of the \emph{CB features}, as there length was modeled in terms of tokens, but here we do so using characters.

\subsection{Contextual Features}

\textbf{Position} (\emph{3 C features}): A sentence on the boundaries of a speaker's segment could contain a reaction to another statement or could provoke a reaction,
which in turn could signal a check-worthy claim.
Thus, we added information about the position of the target sentence in its segment: whether it is first/last, as well as its reciprocal rank in the list of sentences in that segment. 

\textbf{Segment sizes} (\emph{3 C features}): The size of the segment belonging to one speaker might indicate whether the target sentence is part of a long speech, makes a short comment or is in the middle of a discussion with lots of interruptions. The size of the previous and of the next segments is also important in modeling the dialogue flow. Thus, we include three features with the sizes of the previous, the current and the next segments. 

\textbf{Metadata} (\emph{8 C features}): Check-worthy claims often contain mutual accusations between the opponents, as the following example shows (from the 2nd presidential debate):

\begin{footnotesize}
\begin{tabular}{rp{55mm}}
Trump:	& \textbf{Hillary Clinton} attacked those same women and attacked them viciously.	\\
Clinton:	&  They're doing it to try to influence the election for \textbf{Donald Trump}.	\\
\end{tabular}
\end{footnotesize}

Thus, we use a feature that indicates whether the target sentence mentions the name of the opponent, whether the speaker is the moderator, and also who is speaking (3 features). We further use three binary features, indicating whether the target sentence is followed by 
a system message: 
\emph{applause}, \emph{laugh}, or \emph{cross-talk}.\medskip

\subsection{Mixed Features}

The feature groups in this subsection contain a mixture of sentence- and of contextual-level features. For example, if we use a discourse parser to parse the target sentence only, any features we extract from the parse would be sentence-level. However, if we parse an entire segment, we would also have contextual features.

\textbf{Topics} (\emph{300+3 S+C features}): 
Some topics are more likely to be associated with check-worthy claims, and thus we have features modeling the topics in the target sentence as well as in the surrounding context.
We trained a Latent Dirichlet Allocation (LDA) topic model~\cite{blei2003latent} on all political speeches and debates in \emph{The American Presidency Project}\footnote{\url{http://www.presidency.ucsb.edu/debates.php}}
using all US presidential debates in the 2007--2016 period\footnote{\url{https://github.com/paigecm/2016-campaign}}. 
We had 300 topics, and we used the distribution over the topics as a representation for the target sentence. We further modeled the context using cosines with such representations for the previous, the current, and the next segment.

\textbf{Embeddings} (\emph{300+3 S+C features}): We also modeled semantics using word embeddings. 
We used the pre-trained 300-dimensional Google News word embeddings by~\citet{DBLP:journals/corr/MikolovLS13} to compute an average embedding vector for the target sentence, and we used the 300 coordinates of that vector.
We also modeled the context as the cosine between that vector and the vectors for three segments: the previous, the current, and the following one.

\textbf{Discourse} (\emph{2+18 S+C features}): 
We saw above that contradiction can signal the presence of check-worthy claims, and contradiction can be expressed by a discourse relation such as \textsc{Contrast}. As other discourse relations  such as \textsc{Background}, \textsc{Cause}, and \textsc{Elaboration} can also be useful, we used a discourse parser~\cite{jotycodra} to parse the entire segment, and we focused on the relationship between the target sentence and the other sentences in its segment; this gave rise to 18 contextual indicator features. We further analyzed the internal structure of the target sentence ---how many nuclei and how many satellites it contains---, which gave rise to two sentence-level features.

\textbf{Contradictions} (\emph{1+4 S+C features}): Many claims selected for fact-checking contain contradictions to what has been said earlier, as in the example below (from the 3rd presidential debate):

\vspace{3pt}
\begin{footnotesize}
\begin{tabular}{rp{55mm}}
Clinton:	& [\ldots] about a potential nuclear competition in Asia, you said, you know, go ahead, enjoy yourselves, folks.	\\
Trump:		& \textbf{I didn't say} nuclear. \\
\\
\end{tabular}
\end{footnotesize}
We model this by counting the negations in the target sentence as found in a dictionary of negation cues such as \textit{not}, \textit{didn't}, and \textit{never}. 
We further model the context as the number of such cues in the two neighboring sentences from the same segment and the two neighboring segments.

\textbf{Similarity to known positive/negative examples (kNN)} (\emph{2+1 S+C features}): We used three more features inspired by $k$-nearest neighbor (kNN) classification. The first one (sentence-level) uses the maximum over the training sentences of the number of matching words between the testing and the training sentence, which is further multiplied by $-1$ if the latter was not check-worthy. We also used another version of the feature, where we multiplied it by $0$ if the speakers were different (contextual). A third version took as a training set all claims checked by \emph{PolitiFact} (excluding the target sentence).

\section{Experiments and Evaluation}
\label{sec:evaluation}

In this section, we describe our evaluation setup 
and the obtained results.

\subsection{Experimental Setting}
\label{subsec:setting}
We experimented with two learning algorithms. The first one is an SVM classifier with an RBF kernel.\footnote{The RBF kernel worked better than a linear kernel.}  The second one is a deep feed-forward neural network (FNN) with two hidden layers (with 200 and 50 neurons, respectively) and a softmax output unit for the binary classification. We used ReLU \cite{pmlr-v15-glorot11a} as the activation function and we trained the network with Stochastic Gradient Descent \cite{lecun1998gradient}.

The models were trained to classify sentences as positive if \emph{one or more media} had fact-checked a claim inside the target sentence, and negative otherwise. We then used the classifier scores to rank the sentences with respect to \emph{check-worthiness}.\footnote{We also tried using ordinal regression, and SVM-perf, an instantiation of SVM-struct, to directly optimize precision, but none of them yielded improvements.}

\noindent We tuned the parameters and we evaluated the performance using 4-fold cross-validation, using each of the four debates in turn for testing while training on the remaining three ones.

\begin{table}[t]
\small
\centering
\begin{tabular}{l@{}p{0.5cm}p{0.69cm}p{0.5cm}p{0.6cm}p{0.6cm}p{0.6cm}}
    \bf System& \bf MAP & \bf R-Pr & \bf P@5 & \bf P@10 & \bf P@20 & \bf P@50 \\ \hline
    \multicolumn{6}{l}{\bf \emph{Baselines}}\\
     Random      & .164 & .007 & .200 & .125 & .138 & .160\\ 
     TF.IDF	 & .314 & .333& .550 & .475 & .413 & .360\\
     CB Platform & .317 & .349 & .500 & .550 & .488 & .405\\ 
     SVM$_{CBfeat}$ & .360 & .393 & .400 & .425 & .525 & .495\\      
     FNN$_{CBfeat}$ & .357 & .379 & .500 & .550 & .550 & .510\\\hline      
    \multicolumn{6}{l}{\bf \emph{Systems (using all features)}}\\
     SVM$_{All}$     & .395 & .406 & .650 & \bf .725 & .588 & .565\\
     FNN$_{All}$ 	 & \bf .427 & \bf .432 &  \bf .800 & \bf .725 & \bf .713 & \bf .600\\
    \hline
  \end{tabular}
\caption{Evaluation results: our full systems (SVM$_{All}$ and FNN$_{All}$) vs. a number of baselines: random, TF.IDF-based, \emph{ClaimBuster} from the platform, and our two reimplementations thereof.}
\label{table:combination}
\end{table}

For evaluation, we used ranking measures such as \emph{Precision at $k$} ($P@k$) and \emph{Mean Average Precision} (MAP). As Table~\ref{table:per:medium} shows, most media rarely check more than 50 claims per debate. \emph{NPR} and \emph{PolitiFact} are notable exceptions, the former going up to 99; yet, on average there are two claims per sentence, which means that there is no need to fact-check more than 50 sentences even for them. Thus, we report $P@k$ for $k \in \{5, 10, 20, 50\}$.\footnote{Note that while the difference between the P@k metrics (especially between 5 and 10) can be in terms of a few sentences, the deviation between them can seem large}

MAP is the mean of the Average Precision across the four debates. The average precision for a  debate is computed as follows:

 \begin{equation}
 \hbox{AvPrec} = \frac{\sum_{k=1}^{n} (P(k)\times rel(k))} 
 	  {\hbox{number of relevant utterances}} 
 \end{equation}
where $n$ is the number of sentences to rank in the debate, $P(k)$ is the precision at $k$ and $rel(k)=1$ if the utterance at position $k$ is check-worthy, and it is 0 otherwise.

We also measure the recall at the $R$-th position of returned sentences for each debate. $R$ is the number of relevant documents for that debate and the metric is known as $R$-Precision ($R$-Pr).   


\renewcommand{\tabcolsep}{2pt}
\begin{table}[t]
\small
\centering
\begin{tabular}{c@{}l@{}rrrrrr}
 \bf S or C & \bf Feat. Group & \bf MAP & \bf R-Pr & \bf P@5 & \bf P@10 & \bf P@20 & \bf P@50 \\ \hline
 S+C & Embeddings	& .357 & .380 & .450 & .525 & .488 & .495 \\
 S+C & $k$NN		& .313 & .322 & .800 & .725 & .612 & .445 \\
S & Linguistic	& .308 & .333& .450 & .450 & .463 & .430 \\ 
 S & Sentiment	& .260 & .277 &.550 & .400 & .288 & .315 \\ 
 C & Metadata	& .256 & .268 &.350 & .300 & .388 & .370 \\
 S & Length		& .254 & .350&.350 & .375 & .400 & .340 \\ 
 S & NEs	& .236 & .251 & .250 & .275 & .313 & .280 \\   
 S+C & Contradiction	& .222 & .222 &.400 & .275 & .288 & .260 \\
C & Segment size	& .217 & .231 &.100 & .150 & .150 & .245 \\
C & Position	& .212 & .230 &.100 & .075 & .175 & .230 \\
S+C & Discourse	& .205 & .206&.200 & .300 & .325 & .255 \\     
S+C & Topics		& .180 &.178& .000 & .000 & .013 & .085 \\
   \hline
  \end{tabular}
\caption{Performance of each feature group in isolation, when using the FNN system.}
\label{table:features}
\end{table}

\subsection{Evaluation Results}
\label{subsec:results}

Table~\ref{table:combination} shows the performance of our models when using all features described in Section~\ref{sec:model}: see the SVM$_{All}$ and the FNN$_{All}$ rows. 
In order to put the numbers in perspective, we also show the results for five increasingly competitive baselines.  

\noindent First, there is a random baseline, followed by an SVM classifier based on a bag-of-words representation with TF.IDF weights learned on the training data. Then come three versions of the \emph{ClaimBuster} system:
CB-Platform uses scores from the online demo,\footnote{\url{http://idir-server2.uta.edu/claimbuster/demo}} which we accessed on December 20, 2016, and SVM$_{CBfeat}$ and FNN$_{CBfeat}$ are our re-implementations, trained on our dataset.

We can see that all systems perform well above the random baseline. The three versions of ClaimBuster also outperform the TF.IDF baseline on most measures. 
Moreover, our reimplementation of \emph{ClaimBuster} performs better than the online platform in terms of MAP. 
This is expected as their system is trained on a different dataset and it may suffer from testing on slightly out-of-domain data. At the same time, this is reassuring for our implementation of the features, and allows for a more realistic comparison to the \emph{ClaimBuster} system. 

More importantly, both the SVM and the FNN versions of our system consistently outperform all three versions of \emph{ClaimBuster} on all measures.
This means that the extra information coded in our model, mainly more linguistic, structural, and contextual features, has an important contribution to the overall performance. 

We can further see that the neural network model, FNN$_{All}$, clearly outperforms the SVM model: consistently on all measures. As an example, with the precision values achieved by FNN$_{All}$, the system would rank on average 4 positive examples in the list of its top-5 choices, and also 14-15 in the top-20 list. Considering the recall at the first $R$ sentences, we will be able to encounter 43\% of the total number of check-worthy sentences.  This is quite remarkable given the difficulty of the task.

\begin{table}[tbh]
\small
\centering
\begin{tabular}{l@{ }rrrrrr}
  \bf System & \bf MAP & \bf R-Pr & \bf P@5 & \bf P@10 & \bf P@20 & \bf P@50 \\ \hline
  All 	                & .427 &.432& .800 & .725 & .713 & .600\\
  All, no contextual    & .385 & .390 & .550 & .500 & .550 & .540\\
\hline
  Only contextual & .317 & .404 & .725 & .563 & .465 & .465\\
  \it CB Platform & \it .317  & \it .349& \it .500 & \it .550 & \it .488 & \it .405\\ 
\hline
  \end{tabular}
\caption{Impact of the contextual features on the overall performance (FNN system).}
\label{table:context}
\end{table}

\subsection{Individual Feature Types}
\label{subsec:individual}

Table~\ref{table:features} shows the performance of the individual feature groups, which we have described in Section~\ref{sec:model} above, when training using our FNN model, ordered by their decreasing MAP score. We can see that \emph{embeddings} perform best, with MAP of .357 and P@50 of .495. This shows that modeling semantics and the similarity of a sentence against its context is quite important. 

Then come the \emph{kNN} group with MAP of .313 and P@50 of .455. The high performance of this group of features reveals the frequent occurrence of statements that resemble already fact-checked claims. In the case of false claims, this can be seen as an illustration of the essence of our post-truth era, where lies are repeated continuously, in the hope to make them sound true \cite{davies2016age}. 

Then follow two sentence-level features, \emph{linguistic features} and \emph{sentiment}, with MAP of .308 and .260, and P@50 of .430 and .315, respectively; this is on par with previous work, which has focused primarily on similar sentence-level features. 

Then we see the group of contextual features \emph{Metadata} with MAP=.256, and P@50=.370, followed by two sentence-level features: \emph{length} and \emph{named entities}, with MAP of .254 and .236, and P@50 of .340 and .280, respectively. 

At the bottom of the table we find  \emph{position}, a general contextual feature with MAP of .212 and P@50 of .230, followed by \emph{discourse} and \emph{topics}.

\section{Discussion}
\label{sec:discuss}

In this section, we present some in-depth analysis and further discussion.

\subsection{Error Analysis}

We performed error analysis of the decisions made by the Neural Network that uses all available features. Below we present some examples of False Positives (FP) and False Negatives (FN):

\begin{footnotesize}
\begin{tabular}{ll@{}r@{}p{55mm}}
1 &FP & Clinton: & He actually was sued twice by the Justice  Department.\\
2& FP& Clinton: & Five million people lost their homes.\\
3&FP& Clinton: & There's no doubt now that Russia has used cyber attacks against all kinds of organizations in our country, and I am deeply concerned about this.\\
4& FP& Trump: & Your husband signed NAFTA, which was one of the worst things that ever happened to the manufacturing industry.\\
5& FN & Trump: & This is one of the worst deals ever made by any country in history.\\
6& FN & Trump: & Well, nobody was pressing it, nobody was caring much about it.\\
7& FN & Trump: & So Ford is leaving.\\
8& FN & Trump: & It was taken away from her.
\end{tabular}
\end{footnotesize}

The list of false negatives contains sentences that belong to a whole group of annotations and some of them are not check-worthy on their own, e.g., the eighth example. Some of the false negatives, though, need to be fact-checked and our model missed them, e.g., the sixth and the seventh examples. Note also that the fourth and the fifth sentences make the same statement, but they use different wording. On the one hand, the annotators should have labeled both sentences in the same way, and on the other hand, our model should have also labeled them consistently.   

Regarding the false positive examples above, we can see that they could also be potentially interesting for fact-checking as they make some questionable statements. We can conclude that at least some of the false positives of our ranking system could make good candidates for credibility verification, and we demonstrate that the system has successfully extracted common patterns for check-worthiness. Thus, the top-$n$ list will contain mostly sentences that should be fact-checked. Given the discrepancies and the disagreement between the annotations, further cleaning of the dataset might be needed in order to double-check for potentially missing important check-worthy sentences.

\subsection{Effect of Context Modeling}
\label{subsec:context}

Table~\ref{table:context} shows the results when using all features vs. excluding the contextual features vs. using the contextual features only. We can see that the contextual features have a major impact on performance: excluding them yields major drop for all measures, e.g., MAP drops from .427 to .385, and P@5 drops from .800 to .550. The last two rows in the table show that using contextual features only performs about the same as \emph{CB Platform} (which uses no contextual features at all).

\renewcommand{\tabcolsep}{2pt}
\begin{table}[h!]
\centering
\footnotesize
\begin{tabular}{l@{}rrrrrr}
    \bf System & \bf MAP & \bf R-Pr & \bf P@5 & \bf P@10 & \bf P@20 & \bf P@50 \\
    \hline
   \\
    \multicolumn{7}{@{}c}{ \bf   PolitiFact (PF)} \\ 
    \hline
    CB Platform & .154 &.213 & .200 & .300 & .238 & .210\\ 
    \bf NN (train on PF) & .218 & .274 & .450 & .325 & .300 & .270 \\ 
    \bf NN (train on all) & .213 & .246 & .400 & .350 & .375 & .290 \\ 
    \hline
    \\
    \multicolumn{7}{@{}c}{\bf NPR} \\ 
    \hline
   	CB Platform & .144 & .186 &.200 & .225 & .225 & .180\\ 
    \bf NN (train on NPR) & .193 & .216& .550 & .475 & .350 & .255 \\ 
    \bf NN (train on all)& .208 & .250 &.500 & .450 & .375 & .255 \\ 
    \hline
    \\
     \multicolumn{7}{@{}c}{\bf The New York Times (NYT)} \\
     \hline
   	CB Platform & .103 & .250 & .250 & .163 & .135 & .135\\ 
    \bf NN (train on NYT) & .136 &.178& .250 & .225 & .188 & .135 \\ 
    \bf NN (train on all)& .136 &.169& .150 & .200 & .163 & .160\\ 
    \hline
    \\
    \multicolumn{7}{@{}c}{\bf The Guardian (TG)} \\
    \hline
   	CB Platform & .084 &.128& .100 & .100 & .125 & .140\\ 
    \bf NN (train on TG) & .121&.156 & .250 & .225 & .200 & .155 \\ 
    \bf NN (train on all)& .128&.185 & .100 & .150 & .188 & .165 \\ 
    \hline
    \\
     \multicolumn{7}{@{}c}{\bf FactCheck (FC)} \\ 
     \hline
   	CB Platform & .081&.213 & .150 & .125 & .100 & .115\\
    \bf NN (train on FC) & .081&.098 & .050 & .125 & .088 & .085 \\ 
    \bf NN (train on all)& .115&.149 & .100 & .125 & .125 & .140\\ 
    \hline
    \\
    \multicolumn{7}{@{}c}{ \bf   CNN} \\ 
    \hline
    CB Platform & .082 &.096& .150 & .125 & .088 & .085\\ 
    \bf NN (train on CNN) & .079&.076 & .100 & .100 & .100 & .090 \\ 
    \bf NN (train on all)& .095&.087 & .000 & .075 & .088 & .100 \\ 
    \hline
    \\
    \multicolumn{7}{@{}c}{ \bf   Chicago Tribune (CT)} \\ 
    \hline
    CB Platform & .053 &.032& .050 & .050 & .038 & .065\\ 
    \bf NN (train on CT) & .087&.118 & .150 & .150 & .175 & .105 \\ 
    \bf NN (train on all)& .092&.098 & .150 & .075 & .100 & .090 \\ 
    \hline
    \\
    \multicolumn{7}{@{}c}{ \bf   ABC} \\ 
    \hline
    CB Platform & .065 &.066& .150 & .125 & .088 & .080\\ 
    \bf NN (train on ABC) & .059&.068 & .050 & .050 & .100 & .060 \\ 
    \bf NN (train on all)& .088&.090 & .150 & .150 & .113 & .100 \\ 
    \hline
    \\
    \multicolumn{7}{@{}c}{ \bf   Washington Post (WP)} \\ 
    \hline
    CB Platform & .048 &.056& .050 & .075 & .050 & .045\\ 
    \bf NN (train on WP) & .102&.098 & .200 & .175 & .113 & .080 \\ 
    \bf NN (train on all)& .076&.751 & .200 & .100 & .075 & .080 \\ 
    \hline
  \end{tabular}
\caption{Training on the target medium vs. training on all media when testing with respect to a particular medium (using the FNN system).}
\label{table:per:medium:results}
 \end{table}

\subsection{Mimicking Each Particular Source}
\label{subsec:media}

In the experiments above, we have been trying to predict whether a sentence is check-worthy in general, i.e., with respect to at least one source; this is how we trained and this is how we evaluated our models. Here, we want to evaluate how well our models perform at finding sentences that contain claims that would be judged as worthy for fact-checking with respect to each of the individual sources. The purpose is to see to what extent we can make our system potentially useful for a particular medium.

\noindent Another interesting question is whether we should use our generic system or we should retrain with respect to the target medium. 
Table~\ref{table:per:medium:results} shows the results for such a comparison, and it further compares to \emph{CB Platform}. We can see that for all nine media, our model outperforms \emph{CB Platform} in terms of MAP and P@50; this is also true for the other measures in most cases. 

Moreover, we can see that training on all media is generally preferable to training on the target medium only, which shows that they do follow some common principles for selecting what is check-worthy; this means that a general system could serve journalists in all nine, and possibly other, media.\footnote{One exception is Washington Post, where our system performs better when trained on that medium only.} Overall, our model works best on PolitiFact, which is a reputable source for fact checking, as this is their primary expertise. We also do well on NPR, NYT, Guardian, and FactCheck, which is quite encouraging.

\section{Conclusions and Future Work}
\label{sec:conclusion}

We have developed a novel approach for automatically finding check-worthy claims in political debates, which is an understudied problem, despite its importance. Unlike previous work, which has looked primarily at sentences in isolation, here we have focused on the context: relationship between the target statement and the larger context of the debate, interaction between the opponents, and reaction by the moderator and by the public.

Our models have achieved state-of-the-art results, outperforming a strong rivaling system by a margin, while also confirming the importance of the contextual information.
We further compiled, and we are making freely available, a new dataset of manually-annotated claims, 
extracted from the 2016 US presidential and vice-presidential debates, 
which we gathered from nine reputable sources including FactCheck, PolitiFact, CNN, NYT, WP, and NPR.

In future work, we plan to extend our dataset with additional debates, e.g., from other elections, but also with interviews and general discussions.
We would also like to experiment with distant supervision, which would allow us to gather more training data, thus facilitating deep learning.
We further plan to extend our system with finding claims at the sub-sentence level, as well as with automatic fact-checking of the identified claims.

\section*{Acknowledgments}
This research was performed by the Arabic Language Technologies group at Qatar Computing Research Institute, HBKU\@, within the Interactive sYstems for Answer Search project ({\sc Iyas}).

\bibliography{sigproc} 

\begin{thebibliography}{}
\expandafter\ifx\csname natexlab\endcsname\relax\def\natexlab#1{#1}\fi

\bibitem[{Ba et~al.(2016)Ba, Berti-Equille, Shah, and Hammady}]{ba2016vera}
Mouhamadou~Lamine Ba, Laure Berti-Equille, Kushal Shah, and Hossam~M Hammady.
  2016.
\newblock {VERA}: A platform for veracity estimation over web data.
\newblock In {\em Proceedings of the 25th International Conference Companion on
  World Wide Web\/}. Geneva, Switzerland, WWW~'16, pages 159--162.

\bibitem[{Blei et~al.(2003)Blei, Ng, and Jordan}]{blei2003latent}
David~M Blei, Andrew~Y Ng, and Michael~I Jordan. 2003.
\newblock Latent {D}irichlet allocation.
\newblock {\em Journal of Machine Learning Research\/} 3(Jan):993--1022.

\bibitem[{Castillo et~al.(2011)Castillo, Mendoza, and
  Poblete}]{castillo2011information}
Carlos Castillo, Marcelo Mendoza, and Barbara Poblete. 2011.
\newblock Information credibility on {T}witter.
\newblock In {\em Proceedings of the 20th International Conference on World
  Wide Web\/}. New York, NY, USA, WWW~'11, pages 675--684.

\bibitem[{Ciampaglia et~al.(2015)Ciampaglia, Shiralkar, Rocha, Bollen, Menczer,
  and Flammini}]{10.1371journal.pone.0128193}
Giovanni~Luca Ciampaglia, Prashant Shiralkar, Luis~M. Rocha, Johan Bollen,
  Filippo Menczer, and Alessandro Flammini. 2015.
\newblock Computational fact checking from knowledge networks.
\newblock {\em PLOS ONE\/} 10(6):1--13.

\bibitem[{Dagan et~al.(2009)Dagan, Dolan, Magnini, and
  Roth}]{dagan2009recognizing}
Ido Dagan, Bill Dolan, Bernardo Magnini, and Dan Roth. 2009.
\newblock Recognizing textual entailment: Rational, evaluation and approaches.
\newblock {\em Natural Language Engineering\/} 15(4):i--xvii.

\bibitem[{Davies(2016)}]{davies2016age}
William Davies. 2016.
\newblock The age of post-truth politics.
\newblock {\em New York Times\/} 24.

\bibitem[{Ennals et~al.(2010{\natexlab{a}})Ennals, Byler, Agosta, and
  Rosario}]{ennals2010disputed}
Rob Ennals, Dan Byler, John~Mark Agosta, and Barbara Rosario.
  2010{\natexlab{a}}.
\newblock What is disputed on the web?
\newblock In {\em Proceedings of the 4th Workshop on Information
  Credibility\/}. New York, NY, USA, WICOW~'10, pages 67--74.

\bibitem[{Ennals et~al.(2010{\natexlab{b}})Ennals, Trushkowsky, and
  Agosta}]{ennals2010highlighting}
Rob Ennals, Beth Trushkowsky, and John~Mark Agosta. 2010{\natexlab{b}}.
\newblock Highlighting disputed claims on the web.
\newblock In {\em Proceedings of the 19th International Conference on World
  Wide Web\/}. New York, NY, USA, pages 341--350.

\bibitem[{Glorot et~al.(2011)Glorot, Bordes, and Bengio}]{pmlr-v15-glorot11a}
Xavier Glorot, Antoine Bordes, and Yoshua Bengio. 2011.
\newblock Deep sparse rectifier neural networks.
\newblock In {\em Proceedings of the Fourteenth International Conference on
  Artificial Intelligence and Statistics\/}. Fort Lauderdale, FL, USA,
  PMLR~'15, pages 315--323.

\bibitem[{Hardalov et~al.(2016)Hardalov, Koychev, and Nakov}]{Hardalov2016}
Momchil Hardalov, Ivan Koychev, and Preslav Nakov. 2016.
\newblock In search of credible news.
\newblock In {\em Proceedings of the 17th International Conference on
  Artificial Intelligence: Methodology, Systems, and Applications\/}. Varna,
  Bulgaria, AIMSA~'16, pages 172--180.

\bibitem[{Hassan et~al.(2015)Hassan, Li, and Tremayne}]{Hassan:15}
Naeemul Hassan, Chengkai Li, and Mark Tremayne. 2015.
\newblock Detecting check-worthy factual claims in presidential debates.
\newblock In {\em Proceedings of the 24th ACM International Conference on
  Information and Knowledge Management\/}. Melbourne, Australia, CIKM~'15,
  pages 1835--1838.

\bibitem[{Hassan et~al.(2016)Hassan, Tremayne, Arslan, and
  Li}]{Hassan2016ComparingAF}
Naeemul Hassan, Mark Tremayne, Fatma Arslan, and Chengkai Li. 2016.
\newblock Comparing automated factual claim detection against judgments of
  journalism organizations.
\newblock In {\em Computation + Journalism Symposium\/}. Stanford, California,
  USA.

\bibitem[{Hooper(1974)}]{hooper1974assertive}
Joan~B. Hooper. 1974.
\newblock {\em On Assertive Predicates\/}.
\newblock Indiana University Linguistics Club. Indiana University Linguistics
  Club.

\bibitem[{Hyland(1998)}]{hyland1998hedging}
Ken Hyland. 1998.
\newblock {\em Hedging in scientific research articles\/}, volume~54.
\newblock John Benjamins Publishing.

\bibitem[{Joty et~al.(2015)Joty, Carenini, and Ng}]{jotycodra}
Shafiq Joty, Giuseppe Carenini, and Raymond~T. Ng. 2015.
\newblock {CODRA}: A novel discriminative framework for rhetorical analysis.
\newblock {\em Comput. Linguist.\/} 41(3):385--435.

\bibitem[{Karadzhov et~al.(2017{\natexlab{a}})Karadzhov, Gencheva, Nakov, and
  Koychev}]{RANLP2017:clickbait}
Georgi Karadzhov, Pepa Gencheva, Preslav Nakov, and Ivan Koychev.
  2017{\natexlab{a}}.
\newblock We built a fake news \& click-bait filter: What happened next will
  blow your mind!
\newblock In {\em Proceedings of the 2017 International Conference on Recent
  Advances in Natural Language Processing\/}. Varna, Bulgaria, RANLP~'17.

\bibitem[{Karadzhov et~al.(2017{\natexlab{b}})Karadzhov, Nakov, M\`{a}rquez,
  {n}o, and Koychev}]{RANLP2017:factchecking:external}
Georgi Karadzhov, Preslav Nakov, Llu\'{i}s M\`{a}rquez, Alberto Barr\'on-Cede\
  {n}o, and Ivan Koychev. 2017{\natexlab{b}}.
\newblock Fully automated fact checking using external sources.
\newblock In {\em Proceedings of the 2017 International Conference on Recent
  Advances in Natural Language Processing\/}. Varna, Bulgaria, RANLP~'17.

\bibitem[{Karttunen(1971)}]{karttunen1971implicative}
Lauri Karttunen. 1971.
\newblock Implicative verbs.
\newblock {\em Language\/} pages 340--358.

\bibitem[{Le et~al.(2016)Le, Vu, and Blessing}]{le2016towards}
Dieu-Thu Le, Ngoc~Thang Vu, and Andre Blessing. 2016.
\newblock Towards a text analysis system for political debates.
\newblock {\em LaTeCH 2016\/} page 134.

\bibitem[{LeCun et~al.(1998)LeCun, Bottou, Bengio, and
  Haffner}]{lecun1998gradient}
Yann LeCun, Léon Bottou, Yoshua Bengio, and Patrick Haffner. 1998.
\newblock Gradient-based learning applied to document recognition.
\newblock In {\em Proceedings of the IEEE\/}. Anchorage, Alaska, USA, volume~86
  of {\em 1998 IEEE\/}, pages 2278--2324.

\bibitem[{Liu et~al.(2005)Liu, Hu, and Cheng}]{Liu:2005:OOA:1060745.1060797}
Bing Liu, Minqing Hu, and Junsheng Cheng. 2005.
\newblock Opinion observer: Analyzing and comparing opinions on the web.
\newblock In {\em Proceedings of the 14th International Conference on World
  Wide Web\/}. New York, NY, USA, WWW '05, pages 342--351.

\bibitem[{Loper and Bird(2002)}]{Loper02nltk:the}
Edward Loper and Steven Bird. 2002.
\newblock {NLTK}: The natural language toolkit.
\newblock In {\em Proceedings of the ACL-02 Workshop on Effective Tools and
  Methodologies for Teaching Natural Language Processing and Computational
  Linguistics - Volume 1\/}. Philadelphia, Pennsylvania, ETMTNLP '02, pages
  63--70.

\bibitem[{Ma et~al.(2016)Ma, Gao, Mitra, Kwon, Jansen, Wong, and
  Cha}]{ma2016detecting}
Jing Ma, Wei Gao, Prasenjit Mitra, Sejeong Kwon, Bernard~J. Jansen, Kam-Fai
  Wong, and Meeyoung Cha. 2016.
\newblock Detecting rumors from microblogs with recurrent neural networks.
\newblock In {\em Proceedings of the Twenty-Fifth International Joint
  Conference on Artificial Intelligence\/}. New York, New York, USA, IJCAI'16,
  pages 3818--3824.

\bibitem[{Mikolov et~al.(2013)Mikolov, Le, and
  Sutskever}]{DBLP:journals/corr/MikolovLS13}
Tomas Mikolov, Quoc~V. Le, and Ilya Sutskever. 2013.
\newblock Exploiting similarities among languages for machine translation.
\newblock {\em CoRR\/} abs/1309.4168.

\bibitem[{Mohammad and Turney(2013)}]{Mohammad13}
Saif~M. Mohammad and Peter~D. Turney. 2013.
\newblock Crowdsourcing a word-emotion association lexicon 29(3):436--465.

\bibitem[{Nakov et~al.(2017)Nakov, Mihaylova, M\`arquez, Shiroya, and
  Koychev}]{RANLP2017:credibility:trolls}
Preslav Nakov, Tsvetomila Mihaylova, Llu\'is M\`arquez, Yashkumar Shiroya, and
  Ivan Koychev. 2017.
\newblock Do not trust the trolls: Predicting credibility in community question
  answering forums.
\newblock In {\em Proceedings of the 2017 International Conference on Recent
  Advances in Natural Language Processing\/}. Varna, Bulgaria, RANLP~'17.

\bibitem[{Recasens et~al.(2013)Recasens, Danescu-Niculescu-Mizil, and
  Jurafsky}]{Recasens+al:13a}
Marta Recasens, Cristian Danescu-Niculescu-Mizil, and Dan Jurafsky. 2013.
\newblock Linguistic models for analyzing and detecting biased language.
\newblock In {\em Proceedings of the 51st Annual Meeting of the Association for
  Computational Linguistics, Proceedings of the Conference\/}. Sofia, Bulgaria,
  ACL~'13, pages 1650--1659.

\bibitem[{Zubiaga et~al.(2016)Zubiaga, Liakata, Procter, Hoi, and
  Tolmie}]{zubiaga2016analysing}
Arkaitz Zubiaga, Maria Liakata, Rob Procter, Geraldine Wong~Sak Hoi, and Peter
  Tolmie. 2016.
\newblock Analysing how people orient to and spread rumours in social media by
  looking at conversational threads.
\newblock {\em PloS one\/} 11(3):e0150989.

\end{thebibliography}
\bibliographystyle{acl_natbib}
\end{document}